\documentclass[10pt, a4paper]{article}
\usepackage{lrec2022} 
\usepackage{multibib}
\newcites{languageresource}{Language Resources}
\usepackage{graphicx}
\usepackage{tabularx}
\usepackage{soul}
\usepackage{titlesec}
\titleformat{\section}{\normalfont\large\bfseries\center}{\thesection.}{1em}{}
\titleformat{\subsection}{\normalfont\SmallTitleFont\bfseries\raggedright}{\thesubsection.}{1em}{}
\titleformat{\subsubsection}{\normalfont\normalsize\bfseries\raggedright}{\thesubsubsection.}{1em}{}
\renewcommand\thesection{\arabic{section}}
\renewcommand\thesubsection{\thesection.\arabic{subsection}}
\renewcommand\thesubsubsection{\thesubsection.\arabic{subsubsection}}

\usepackage{epstopdf}
\usepackage[utf8]{inputenc}

\usepackage{hyperref}
\usepackage{xstring}

\usepackage{color}
\usepackage[table]{xcolor}

\definecolor[named]{utugray}{rgb}{.85,.85,.85} 

\title{Out-of-Domain Evaluation of Finnish Dependency Parsing  }

\name{Jenna Kanerva and Filip Ginter} 

\address{TurkuNLP, Department of Computing \\
         University of Turku, Finland \\
         \{jmnybl, figint\}@utu.fi\\}

\abstract{
The prevailing practice in the academia is to evaluate the model performance on in-domain evaluation data typically set aside from the training corpus. However, in many real world applications the data on which the model is applied may very substantially differ from the characteristics of the training data. In this paper, we focus on Finnish out-of-domain parsing by introducing a novel UD Finnish-OOD out-of-domain treebank including five very distinct data sources (web documents, clinical, online discussions, tweets, and poetry), and a total of 19,382 syntactic words in 2,122 sentences released under the Universal Dependencies framework. Together with the new treebank, we present extensive out-of-domain parsing evaluation utilizing the available section-level information from three different Finnish UD treebanks (TDT, PUD, OOD). Compared to the previously existing treebanks, the new Finnish-OOD is shown include sections more challenging for the general parser, creating an interesting evaluation setting and yielding valuable information for those applying the parser outside of its training domain.  \\ \newline  }

\begin{document}

\maketitleabstract

\section{Introduction}

During software development and model performance evaluation, the prevailing practice in the academia is to evaluate the model performance on an in-domain dataset. This typically means that the model is evaluated on a test section set aside from the training corpus, therefore the test dataset sharing the same properties as the data used to train the model. However, in many real world applications this may not be the case. The data on which the model is applied in its actual use in downstream applications may in practice very substantially differ from the characteristics of the training data.

In this paper, we focus on Finnish dependency parsing in the Universal Dependencies (UD) scheme. When both trained and tested on the UD dataset, the state of the art is approaching human performance \cite{virtanen2019multilingual}. Consequently, the Finnish parser is in active use in the academia as well as in the commercial industry, and applied in numerous downstream tasks and domains as a text normalization and pre-processing component. In some cases, however, these application domains substantially differ in their characteristics from the training corpus and there is no hard evidence as to the effect on the parser performance.

To address this question, we selected and annotated a new treebank meant solely for out-of-domain evaluation of the models trained on the UD Finnish-TDT dataset. This new UD Finnish-OOD treebank allows us to quantify the parsing performance in various downstream applications, and to better understand the limits of generalization exhibited by the most recent dependency parsing methodology. In addition to introducing the new dataset, we carry out several Finnish out-of-domain parsing experiments, where in addition to the presented Finnish-OOD treebank, we use the existing section-level metadata in order to carry out section level performance evaluation. We show the new Finnish-OOD treebank being more challenging compared to the different sections existing in the current treebanks available for Finnish in the UD collection.

\section{Data Sources}

The three UD treebanks presently available for Finnish, namely Finnish-TDT~\cite{haverinen2013tdt,pyysalo2015udfinnish}, Finnish-PUD~\cite{zeman-etal-2017-conll}, and Finnish-FTB~\citelanguageresource{finntreebank1}, represent 6 different text genres based on the UD genre classification: blogs, fiction, grammar examples, legal, news and Wikipedia. 

The UD Finnish-TDT is a general domain treebank containing 202,453 syntactic words (15,136 sentences) from 10 different text sources: Wikipedia articles, online fiction, JRC-Acquis legislation, popular online blogs, EuroParl speeches, grammar examples, Wikinews, university news, economy news, and student magazine articles. The treebank is divided into training, development and test set, the training set of the TDT treebank being the primary training data used throughout all experiments reported in this study. The UD Finnish-PUD is the Finnish part of the parallel UD treebank collection annotating the same underlying text translated for multiple languages. It includes 15,317 words (1,000 sentences) from two text sources: Wikipedia and news. The Finnish-PUD is used as external test set in this study. The UD Finnish-FTB is a treebank of grammar book examples annotated as a separate effort, independently of the other two Finnish UD treebanks and its annotation is not compatible in many important details with the abovementioned treebanks. Since these incompatibilites would mask any interesting differences, we do not use the FTB treebank in this study. Nevertheless, the \emph{Grammar examples} section of the  TDT treebank is in fact sampled from the FTB text material, and therefore the FTB treebank text domain is represented in our experiments.


In order to build the new out-of-domain treebank for Finnish UD parsing we consider four text genres defined in UD v2.8 genre classification (see e.g\ \citelanguageresource{udv2.8}, \cite{muller2021howuniversal}) but absent from the previously available Finnish treebanks: medical, poetry, social and web. Under these four genres, we include documents from five different text sources: (1) clinical nursing narratives of hospital patients for the medical domain, (2) web documents manually identified to contain poems or song lyrics for poetry, (3) discussion forum messages and (4) tweets for the social domain, and (5) randomly sampled documents from a general internet crawl for web. Of these, especially the nursing narratives, poetry, and tweets differ very substantially from any text in the training data. Yet, in particular the clinical domain and tweets represent a typical application domain for the Finnish parser.

In the following, we describe the data collection and cleaning procedure for each data source separately.

\subsection{Medical -- Clinical Nursing Narratives}
\label{sec:clinical-annotation}

In clinical nursing narratives the patient's visit in the hospital is recorded in a free text narrative, where the document is amended by nurses throughout their shifts to describe the patient's condition, treatments and status development during the stay in the hospital. These nursing records are meant for medical professionals to help in clinical decision making, and due to the fact that the records are targeted to professionals, the text is heavy on special terminology. Additionally, the nature of nursing narratives substantially differ from general language use, nursing narratives oftentimes including only critical facts expressed in a sentence without a main verb rather than carefully edited sentences. \newcite{laippala2014sp} described the key characteristics in nursing narratives including frequent misspellings, abbreviations, domain terminology, telegraphic writing style and non-standard syntactic structures.

\newcite{suominen2009nursing} collected a corpus of nursing narratives of Finnish intensive care unit patients in the Turku University Hospital during years 2005-2006. In this work the nursing narratives of selected patients (those whose stay in the intensive care unit was at least 5 days) where extracted from the hospital's electronic patient records. Due to the obvious considerations on personal health data, this corpus is not available online. However, later on, a small section of this corpus consisting of 8 full narratives was manually anonymized and made openly available with annotation in the Stanford Dependencies (SD) scheme \cite{haverinen09parsing,haverinen-etal-2010-dependency}. While the dependency relations were manually annotated in this clinical Finnish treebank, the segmentation, morphology and lemma annotation layers were only automatically predicted. In this work, we sample two full narratives (939 sentences) from the original clinical treebank, and manually re-annotate them into UD scheme, this time including manual annotation of all layers, i.e. morphological tags, lemmas, and syntax.

The original clinical treebank data is available only with automatic segmentation, and since we do not have access to the original corpus used to obtain the anonymized records, the information on original text is lost. In order to include at least a minimal support towards testing segmentation models on the medical data, we apply manual detokenization, where e.g.\ punctuation markers are reconnected with the previous tokens when applicable, and thus the text is ``corrected'' to reflect orthographic standards in the original corpus. Misspellings and other segmentation issues that we could assume not to be introduced by the original tokenizer were left as-is. By doing this, we recognize the issue of the detokenized data not fully reflecting the possible variation of misspellings in cases where it was unclear whether the error was introduced by the original writer or the automatic tokenizer.


\subsection{Poetry -- Poems and Song Lyrics}

In our poetry subsection, we rely on web documents manually identified to include poems or song lyrics. These documents are drawn from the FinCORE corpus~\cite{laippala-etal-2019-toward}, where a random sample of Finnish web crawled documents are manually labeled for their text register, using 8 top-level labels (narrative, opinion, informational description, interactive discussion, how-to/instructional, informational persuasion, lyrical, and spoken) and several subcategories. We extract all documents manually labeled as lyrical in the FinCORE corpus, denoting a top-level category including both poems and song lyrics. At the time of data collection, we were able to identify 6 documents (144 sentences) with the lyrical label.


\subsection{Social --- Tweets}

The Finnish tweets were downloaded between years 2016-2018 using the Twitter streaming API \footnote{We used the Tweepy Library \url{https://github.com/tweepy/tweepy}.}. During downloading, we keep only tweets labeled as Finnish by the Twitter's language recognition (available in the tweet json). However, we observed the downloaded dataset to include a large number of tweets incorrectly labeled as Finnish, and therefore we manually identified all Finnish tweets from a small sample of 1250 tweets randomly selected among all downloaded tweets. This manual curation step discarded over 50\% of all sampled tweets, indicating the language identification labels not being accurate enough for selecting Finnish tweets.\footnote{All sampled tweets with their manually annotated labels are available at \url{https://github.com/TurkuNLP/finnish-tweets-lang-identification} for any later experiments on language identification of Finnish tweets.} Finally, 130 randomly sampled tweets from the curated dataset proceeded into the manual morphosyntactic annotation.

Likely due to change in Twitter character limits in 2017, the main text field in the downloaded tweet json sometimes contains a truncated version of the tweet. Similarly for retweets, the main text field includes a retweet marker (\texttt{RT @USERNAME:}), and the actual tweet can become truncated. In both cases, we always extract the full tweet text rather than the truncated one. This also has the property of not including the retweet marker as part of the extracted text, but retaining the information in the corpus metadata. This strives to mimic the textual content of a tweet as the user would see it through the online interface.

\subsection{Social -- Discussion Forum Messages}

The second subset of social network data is gathered from the Suomi24 corpus\footnote{\url{http://urn.fi/urn:nbn:fi:lb-2020021802}}, containing all messages posted in the Finnish Suomi24 online discussion forum between years 2001 and 2017. Historically, it has been one of the largest social network forums in Finland and covers a broad range of discussion topics including language with wide range of different writing styles and formality. From this dataset, we randomly sample 51 different messages for manual annotation, where messages may be anything between quick reactions to previous messages to longer posts on any number of different topics.

\subsection{Web -- Random Sample of the Internet Crawl}

For the web domain, we take a random sample of 30 documents from the Finnish Internet Parsebank~\cite{luotolahti2015parsebanks}. Five documents manually determined to be machine translated, thus, including many incomprehensible sentences, were replaced with new documents during sampling. Due to many web documents being quite long, each document was truncated after 25 sentences in order to avoid overly long documents biasing the web data towards particular topics. Furthermore, unnatural repetition appearing in some documents was removed (e.g.\ repeating quotations blocks) to avoid artificially skewing the evaluation statistics, and in these cases, more sentences were taken from the same document until the 25 sentence limit was reached.


\section{Treebank Annotation}

The data was annotated by a single annotator with a long-term experience in Finnish UD treebanking and the sole maintainer of the UD Finnish-TDT corpus. In the annotation, the Universal Dependencies guidelines were used as adapted in the Finnish-TDT corpus, thus making the corpus suitable for out-of-domain experiments especially for models trained with the Finnish-TDT treebank and making the new corpus fully compatible with UD Finnish-TDT in the numerous analysis choices and guideline interpretations. The dataset is natively annotated into the UD scheme, including fully manual analysis of all relevant layers (segmentation, morphology, lemmas and dependency syntax).

While some of the new text sources can be quite straightforwardly annotated using the general guidelines, some of the domains need domain-specific choices, as there are no established prior guidelines for some of the constructions. By far the most difficult domains in this work were poetry due to its specialities in sentence segmentation, and tweets due to including tokens limited to social media texts (e.g.\ hashtags and mentions), not appearing in the Finnish-TDT treebank. In addition to these, the medical domain posed interesting challenges in its specific medical terminology, while discussion forum messages and web documents did not substantially differentiate from the general domain texts in terms of annotation, and therefore, did not require adaptations to the general guidelines.

Next, we will discuss the annotation process separately for the poetry, tweets and clinical nursing narratives, as well as discuss the most relevant related work supporting the annotation decisions made during the annotation.

\subsection{Clinical Nursing Narratives}
While some of the medical terms used in the clinical nursing narratives are easily understandable to readers without professional medical knowledge, some terms require domain-specific understanding in order to correctly determine their meaning. Clearly, the annotation of morphological features and dependency relations for such terms is difficult for a person working outside the domain. While many of such terms are available in different medical dictionaries, especially highly abbreviated versions of medical terms are oftentimes difficult to find. In order to support the corpus annotation, we start the annotation process by translating all domain-specific terms into a general language with the help of a trained nurse. These translations are included as additional annotation in the MISC field of the CoNLL-U file, where the translations could be provided on word-to-word basis (\texttt{Gen=Translation}). An informative description of a concept is included instead in the MISC field in cases where a word-to-word translation is not feasible (\texttt{Gen\textunderscore desc=Description}).


In general, the medical domain is quite rare in UD treebanks, in addition to ours, only 6 UD treebanks are reported as including medical texts. In fact, in all of these 6 treebanks (Czech-CAC~\cite{raab2008czech}, French-Sequoia~\cite{candito2012frenchseq}, Kiche-IU~\cite{tyers_2021}, Persian-Seraji~\cite{seraji2016universal}, Romanian-RRT~\cite{mititelu2018modern}, and Romanian-SiMoNERo~\cite{mititelu2020romanian}), the medical texts are reported to be based on scientific or technical writings from the field of medicine, thus being carefully edited, official publications. In contrast, the clinical nursing narratives used in our corpus are quickly drafted notes written to other professionals and not meant to be publicly shared, making the nature of our medical texts very different from other UD treebanks. However, after dealing with the terminology, the rest of the annotation work was quite straightforward.


\subsection{Poetry}

While annotating texts from the poetry genre, one feature clearly standing out was the usage of capitalization and line breaks to articulate the layout of the text (indicating rhythm) rather than following standard structure of dividing text into paragraphs and sentences. In some documents, this resulted in having long text passages without punctuation characters indicating the standard sentence or clause structures.

Similar to medical, poetry is also among one of the rarest genres in UD. In addition to our treebank only 6 datasets are reported as including it: Belarusian-HSE~\footnote{\url{https://github.com/UniversalDependencies/UD_Belarusian-HSE}}, Breton-KEB~\cite{tyers_2021}, Latin-UDante~\cite{cecchini2020udante}, Old French-SRCMF~\cite{stein2013syntactic}, Romanian-Nonstandard~\cite{maranduc2017romania}, and Russian-Taiga~\cite{shavrina2017taiga}. While the segmentation of poetry texts is not explicitly mentioned in the studies or UD specifications, we follow a similar principle that seems to be the consensus in other UD treebanks, based on our understanding of the examples available in the papers, as well as inspecting annotated sentences in the released datasets.

We segment the texts into sentences following the existence of the sentence-final punctuation rather than capitalization or single line breaks, as oftentimes the text after a single newline was evidently a continuum of the previous sentence (fitting e.g.\ dependency relations \texttt{obl}, \texttt{advcl},  \texttt{acl}, or \texttt{conj}). In such cases where the sentence continuation was semantically ambiguous (full stop could have been easily used to break the sentence), these segments are connected with the parataxis relation marking for side-by-side clauses without coordination, subordination or argument relation. An exception, where we follow line breaks rather than punctuation, is made with double newlines (indicating paragraph or stanza boundary), where the sentence boundary is annotated even without an explicit sentence-final punctuation.







\subsection{Tweets}
\begin{figure*}[t]
   
   Peliriippuvuudesta tuli viimein \#oikea \#sairaus -- WHO lis\"{a}si virallisiin tautiluokituksiin \#peliaddiktio \\ \#peliriippuvuus \#WHO \#tautiluokitus https://t.co/P8wSQZzW45
    
    \vspace{3mm}
    \textit{Gaming disorder finally became a \#real \#disease -- WHO added (it) to the official classification of diseases \\ \#gamingdisorder \#gamingaddiction \#WHO \#classificationofdiseases https://t.co/P8wSQZzW45}
   
    \caption{An example of a typical tweet including hashtags both as replacing normal, content-bearing words (\#oikea/\#real and \#sairaus/\#disease) as well as listing topical keywords at the end.
    }
    \label{fig:tweet}
\end{figure*}

Tweets include several characteristics rather unique to limited social media channels, the most common being mentions (\texttt{@username}) and hashtags (\texttt{\#hashtag}), while also URLs and emoticons are substantially more frequent in tweets than in many other genres in the Finnish treebanks. Also, due to the character limits in social media platforms, tweets are rather short documents typically including only one or two short sentences. Likely due to this reason, in many treebanks including Twitter data, tweets are considered to be single sentence units and further sentence splitting is not applied (see e.g.\ the data releases of Italian-PoSTWITA \cite{sanguinetti2017annotating}, Italian-TWITTIRO \cite{cignarella2019presenting}, or Tweebank by \newcite{liu2018parsing}). However, based on manual annotation 35\% of the tweets in our sample include more than one sentence, 72\% of sentences in these multi-sentence tweets containing a predicate in the main clause, thus indicating the individual sentences more often being real sentence-like units rather than short noun phrases. As the CoNLL-U format supports indicating document structure as metadata, we do not want to artificially analyse tweets as single sentences, when similar text passages in any other genre would be segmented into multiple sentences. Therefore, we consider a tweet as a small document which is further segmented into sentences as necessary. However, special tokens (mentions and hashtags) as well as plain interjections (e.g.\ \emph{Wonderful!}) in the beginning or end of the tweet are kept together with the corresponding sentence. Regarding interjections, a similar approach is applied also in the Finnish-TDT treebank, thus not making deviation to the original annotation scheme. Regarding token segmentation, we treat mentions and hashtags as single tokens, where the special characters (\texttt{@} or \texttt{\#}) are simply part of the main token. Otherwise standard tokenization guidelines are applied.

While there are several studies involving UD annotation on tweets (see e.g.\ \newcite{sanguinetti2017annotating}, \newcite{liu2018parsing}, \newcite{bhat2018universal}, and \newcite{blodgett2018twitter}), there does not seem to be a clear consensus regarding the annotation of tokens specific to Twitter or other social media platforms. Mentions are usernames appearing typically at the beginning of the sentence to mark dialogue participant in addressed speech, or occasionally replacing a normal content-bearing word in the sentence, usually when referring to an entity which would otherwise be a proper name (e.g.\ person or company name). Hashtags have a similar distinction, where most of the hashtags are used as a list of topical keywords appearing in the beginning or at the end of the sentence, however, some can be used as normal content-bearing words to replace any normal token in the sentence. In Figure~\ref{fig:tweet} we illustrate a typical tweet taken from the corpus.


While \newcite{sanguinetti2017annotating} and \newcite{bhat2018universal} annotated mentions with \texttt{SYM} part-of-speech tag, \newcite{liu2018parsing} used \texttt{PROPN}, however, all agreeing of using \texttt{vocative} dependency relation for those mentions appearing in the beginning of the tweet to address the dialogue participant. As mentions are references to Twitter usernames and thus can be considered as proper names, we opted for labeling all mentions with \texttt{PROPN} on the part-of-speech level, while the syntactic relation depends on how the token is used. For mentions addressing the dialogue participant we follow the other treebanks by annotating them with the \texttt{vocative} dependency relation, while those used as content-bearing words are annotated with their corresponding function in the sentence (e.g.\ subject or object).

Hashtags are annotated in various ways in the released treebanks. \newcite{sanguinetti2017annotating} and \newcite{bhat2018universal} analyse all hashtags as symbols (\texttt{SYM}), whereas \newcite{liu2018parsing} uses \texttt{X} for topical hashtags, while annotating content-bearing hashtags as any normal tokens. In terms of relations, both \newcite{sanguinetti2017annotating} and \newcite{liu2018parsing} use the corresponding relations in the sentence for content-bearing hashtags, while \newcite{bhat2018universal} and \newcite{blodgett2018twitter} do not distinguish content-bearing hashtags from the topical ones. The topical hashtags are annotated as \texttt{parataxis} \cite{sanguinetti2017annotating,blodgett2018twitter}, or \texttt{discourse} \cite{liu2018parsing,bhat2018universal}. We opted for analysing hashtags with their corresponding part-of-speech tags when the token is an actual Finnish word (i.e.\ \texttt{\#beautiful} would be an adjective and \texttt{\#forest} a noun). However, in some cases giving a real part-of-speech analyse for a hashtag is not meaningful, this would be the case for example with foreign words or tokens artificially joining several words together (e.g.\ \texttt{\#thisisbeautiful}). For these, the \texttt{X} part-of-speech tag is used in the same manner as would be done with similar regular tokens as well. In the relation annotation, we annotate topical hashtags with the \texttt{discourse} relation, while content-bearing hashtags receive annotation regarding its real syntactic function in the sentence.

Due to the choices done during the text preprocessing, retweet markers often appearing in Twitter corpora (such as \texttt{RT} in the beginning of a tweet), do not appear in our data. Regarding URLs and emoticons quite frequently occurring in the corpus, we follow the general Finnish-TDT annotation standards, where both are annotated as symbols (\texttt{SYM}) in the part-of-speech level. While in Finnish-TDT emoticons are always attached with the \texttt{discourse} relation to the sentence root, the relation and attachment of URLs depend on the sentence context. However, most of the URLs appearing in tweets are sentence-final referential items, which do not hold any content-bearing function, we use the same \texttt{discourse} relation for such URLs as well.

\section{Treebank Statistics}

The statistics of the Finnish-OOD corpus are summarized in Table~\ref{tab:data-statistics}, where the section-specific document, sentence and syntactic word counts are plotted together with the two other corpora annotated using the same guidelines and used later in the parsing experiments, Finnish-TDT and Finnish-PUD. The total size of the Finnish-OOD corpus is 19,382 syntactic words (2,122 sentences), where \emph{syntactic word} is the basic element of syntactic annotation in Universal Dependencies. The different subsections vary in size between 2,005 (poetry) and 6,906 words (web documents). The whole corpus is released as test data only.

For comparison, among the 217 test sets in the present Universal Dependencies release 2.9, the average length is 17,946 words and median length is 11,385 words. This makes the Finnish-OOD with its 19,382 words an average UD test set in length, in fact considerably above the median length, ranking 53th out of 217. In terms of full UD treebanks (not only their test sets), Finnish-OOD still contains more total words than 81 of the 217 UD treebanks.



\begin{table*}[]
    \centering
    \begin{tabular}{lrrr|rrr|rrr}
        & \multicolumn{3}{c}{\textbf{Train}} & \multicolumn{3}{c}{\textbf{Dev}} & \multicolumn{3}{c}{\textbf{Test}}  \\ 
       \textbf{Section}  & \textbf{Doc.} & \textbf{Sent.} & \textbf{Words} & \textbf{Doc.} & \textbf{Sent.} & \textbf{Words} & \textbf{Doc.} & \textbf{Sent.} & \textbf{Words}  \\ 
       \rowcolor{utugray}
       \multicolumn{10}{l}{Finnish-TDT} \\
       Wikipedia & 160 & 1,799 & 25,109 & 20 & 200 & 2,890 & 20 & 270 & 3,936 \\
       Fiction & 51 & 2,202 & 26,342 & 7 & 221 & 2,785 & 7 & 316 & 3,732 \\
       Legal & 23 & 914 & 19,130 & 3 & 85 & 1,938 & 3 & 142 & 2,892 \\
       Blogs & 61 & 1,356 & 16,773 & 8 & 259 & 3,348 & 8 & 166 & 2,219 \\
       EuroParl & 64 & 872 & 16,298 & 8 & 94 & 1,674 & 8 & 116 & 1,986 \\
       Grammar examples & --- & 1,601 & 13,608 & --- & 200 & 1,623 & --- & 201 & 1,771 \\
       Wikinews & 80 & 921 & 11,953 & 10 & 92 & 1,086 & 10 & 107 & 1,256 \\
       University news & 40 & 765 & 10,644 & 5 & 86 & 1,342 & 5 & 91 & 1,243 \\
       Economy news & 40 & 854 & 10,499 & 5 & 63 & 821 & 5 & 85 & 1,136 \\
       Student magazines & 19 & 933 & 12,668 & 2 & 64 & 823 & 2 & 61 & 928 \\
        \textbf{Total} & --- & \textbf{12,217} & \textbf{163,024} & --- & \textbf{1,364} & \textbf{18,330}  & --- & \textbf{1,555} & \textbf{21,099}  \\
        \rowcolor{utugray}
        \multicolumn{10}{l}{Finnish-PUD} \\
        Wikipedia &&&&&&    & 251 & 625   & 9,901   \\
        News    &&&&&&     & 146 & 375   & 5,916    \\
        \textbf{Total}  &&&&&&      & \textbf{397} & \textbf{1,000} & \textbf{15,817} \\
        \rowcolor{utugray}
        \multicolumn{10}{l}{Finnish-OOD} \\
        Web documents     &&&&&&   & 30  & 584   & 6,906  \\
        Clinical  &&&&&&  & 2   & 939   & 5,330 \\ 
        Online discussions &&&&&& & 51  & 263   & 3,071  \\
        Tweets  &&&&&&   & 130 & 192   & 2,070 \\
        Poetry  &&&&&&   & 6   & 144   & 2,005  \\
        \textbf{Total}  &&&&&&    & \textbf{218} & \textbf{2,122} & \textbf{19,382} \\ 
        
    \end{tabular}
    \caption{Section-specific statistics for Finnish TDT, PUD and OOD treebanks in terms of document, sentence and token counts. Sections in each treebank are sorted in descending order based on the test set token count.}
    \label{tab:data-statistics}
\end{table*}

\section{Out-of-domain Parsing}

In this section we report on dependency parsing experiments, where the parser trained on the Finnish-TDT treebank is tested both on its in-domain data (Finnish-TDT) and out-of-domain data using the newly introduced Finnish-OOD and the existing Finnish-PUD datasets. First, we measure off-the-shelf parsing performance on these datasets in order to report baseline performance directly comparable to other studies, and later perform several detailed section-wise analyses. Additionally, since the Finnish-TDT treebank preserves metadata about the original text sources, we also carry out ``leave section out'' experiments across the 10 sections of the Finnish-TDT treebank, obtaining further out-of-domain parsing experiments. These allow us to gauge the benefit of the new dataset compared to what was available previously.


The parsing experiments are carried out using the Turku Neural Parser Pipeline~\cite{udst:turkunlp}, which is a full parsing pipeline with parsing accuracy at the level of present state-of-the-art for UD Finnish parsing. Updated from its original release, the current pipeline consist of a segmentation module based on the UDPipe implementation~\cite{udpipe:2017}, custom part-of-speech and morphological feature tagger including separate POS and feature classification layers on top of shared pre-trained embeddings, graph-based bi-affine parser of \cite{dozat2017stanford} based on its implementation in Diaparser\footnote{\url{https://github.com/Unipisa/diaparser}}, and a sequence-to-sequence lemmatizer by \newcite{universallemmatizer}. Out of these four components, the tagger and parser utilize the pre-trained FinBERT language model by \cite{virtanen2019multilingual}, while the segmenter and lemmatizer modules do not rely on any pre-training.

\begin{table*}[]
    \centering
    \begin{tabular}{lccccccccccc}
    \rowcolor{utugray}
       \textbf{Treebank} & \textbf{Tokens} & \textbf{Sent.} & \textbf{Words} & \textbf{UPOS}  & \textbf{UFeats} & \textbf{Lemmas} & \textbf{UAS}   & \textbf{LAS} \\ 
        TDT & 99.6 & 87.2 & 99.6 & 97.9 & 96.7 & 95.8 & 93.0 & 91.0\\
        PUD & 99.6 & 91.3 & 99.6 & 98.0 & 97.1 & 95.3 & 94.0 & 92.1\\
        OOD & 97.6 & 65.5 & 97.5 & 92.5 & 91.9 & 91.1 & 81.6 & 77.5     
    \end{tabular}
    \caption{Baseline parsing experiments for the parser trained on the TDT data, and tested on its own test set (TDT) as well as two external test sets (PUD, OOD).}
    \label{tab:baseline-exp}
\end{table*}

In Table~\ref{tab:baseline-exp} we report the baseline experiments, where the parser trained on the full TDT corpus training set is evaluated on its own test set (TDT) as well as the two external test sets (PUD and OOD). When applying the model to the PUD dataset, the parsing performance does not decrease, the LAS performance actually being +1pp higher compared to the original TDT test set. Similar observations are reported in multiple other studies as well (see e.g.\ \newcite{zeman-etal-2017-conll}), suggesting the PUD test set being easier compared to the TDT test set. Additionally, one must take into account the fact that although we treat PUD as external, separately constructed treebank, the sections included in PUD have a major domain overlap between those in TDT (namely Wikipedia for PUD Wikipedia and Wikinews, economy news and university news for PUD news). Therefore, the PUD dataset cannot be considered as out-of-domain data for TDT trained models (and was, in fact, never meant to be an out-of-domain test set in the first place). On the contrary to PUD, the parsing performance drastically decreases on the newly introduced Finnish-OOD dataset, LAS decreasing over 13pp from 91.00 to 77.50.



Next, we set out to study this further by breaking down the data section-by-section in each of the three treebanks, and carrying out the ``leave section out'' experiments also across the 10 different sections of the Finnish-TDT treebank. In these ``leave section out'' experiments, the trained model has never seen data from the particular section during model training, thus demonstrating the out-of-domain parsing performance on the TDT treebank also. In respect of the two PUD sections (Wikipedia and news), we report numbers when leaving all corresponding TDT sections out during training, while in OOD the model is trained on full TDT data as there is no domain overlap between the treebank sections. 

\begin{table*}[]
    \centering
    \includegraphics[width=\textwidth]{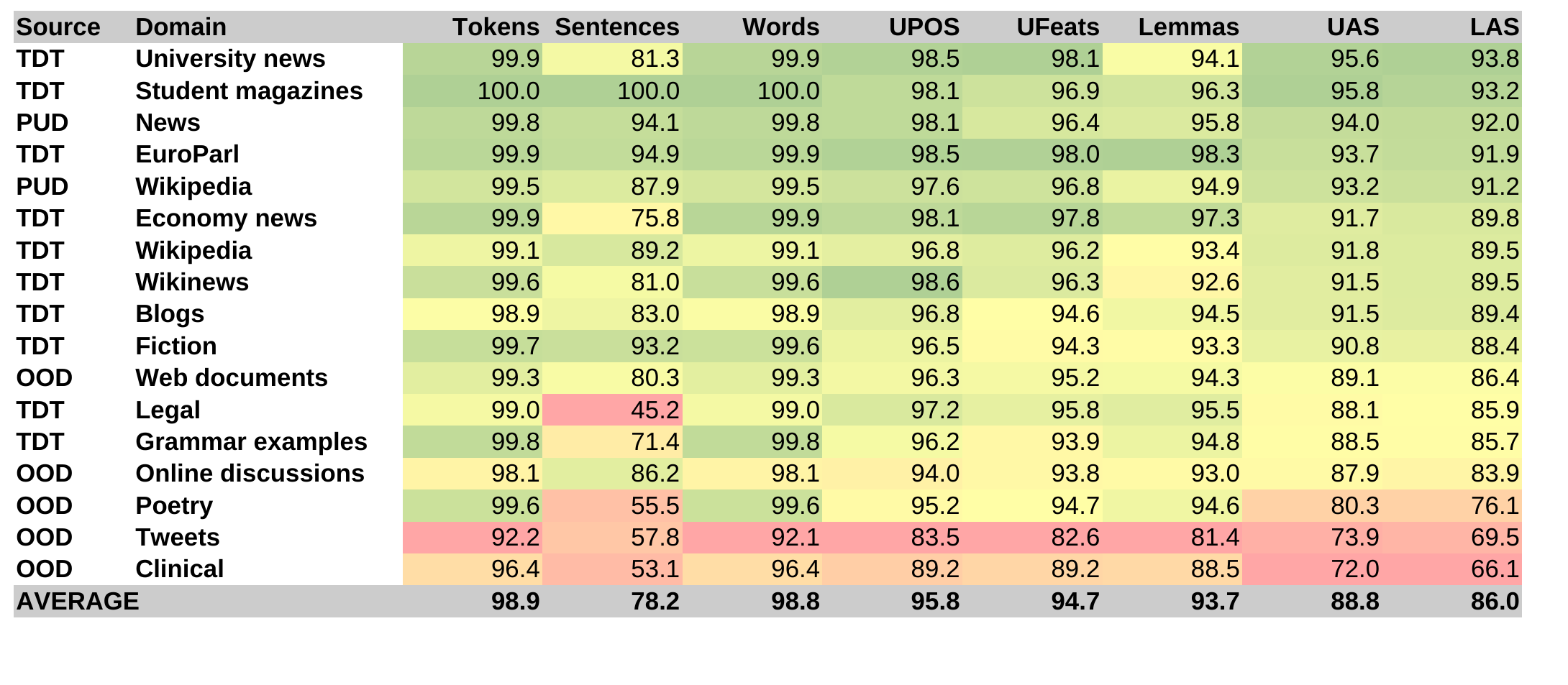}
    \caption{Out-of-domain parsing performance of a model trained on UD Finnish-TDT and tested on all available test sets. All tests are out-of-domain, i.e.\ if the test set originates from the TDT treebank, the relevant section is removed from the training data. Similarly, for the PUD test set sections, the corresponding domain was removed from the training data. The results are sorted by LAS and color-coded by difference from the average.}
    \label{tab:all-sections}
\end{table*}

The section-wise results are shown in Table~\ref{tab:all-sections}. In terms of the OOD sections (web documents, clinical, online discussions, tweets, and poetry), quite unsurprisingly the two best performing out-of-domain sections are web documents and online discussions in terms of parsing accuracy (LAS metric), those sections not very substantially differing from the general data in terms of data annotation, and thus likely closest to the genres seen during the model training as well. In addition to the treebank data, the pre-trained FinBERT model used as starting point in parser fine-tuning, was trained on a large collection of web and discussion forum data. During pre-training, the FinBERT language model used 3.3B tokens of Finnish including discussion forum data (52\%), web crawl (33\%), and news (15\%). Therefore, although these two genres are out-of-domain in terms of parser training, the parser was exposed to these genres through language model pre-training.

On the other end of the scale in terms of LAS is the clinical domain text, nearly 26pp below the in-domain performance, an accuracy level which is likely too low for practical applications. The parsing accuracy on tweets is about 20pp below the in-domain performance, also a very substantial drop. Other measures, such as the accuracy of POS and morphological tagging, and lemmatization, on the other hand, do not exhibit nearly as substantial drop as the syntactic tree accuracy. Especially lemmatization, which is an important step in search and indexing type of applications, sees a comparatively moderate absolute drop in performance across the various OOD subdomains.

When comparing the different sections from all three treebanks, it's clear that the sections selected for the Finnish-OOD are in general more difficult that the ones present in TDT and PUD even when evaluated in ``leave section out'' manner. With the exception of OOD web documents having higher LAS than TDT legal and grammar examples, the OOD sections locate to the bottom of the table when sorted in terms of LAS in descending order.


\begin{table*}[t]
\centering
\includegraphics[width=\textwidth]{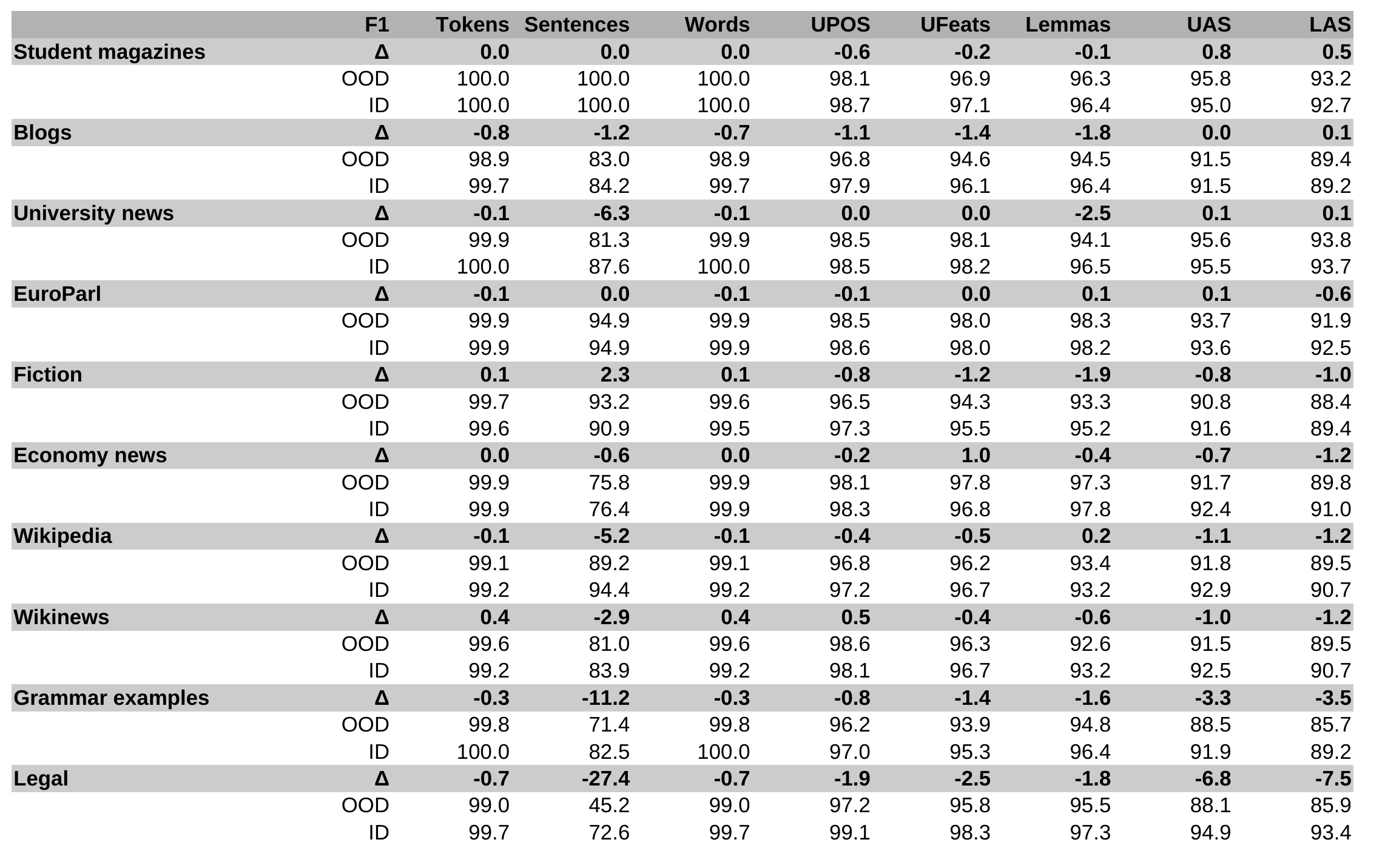}
\caption{Parsing performance on the various sections of the UD Finnish TDT treebank. OOD refers to an out-of-domain run, where the section is removed from the training data, while ID refers to an in-domain run, where the section is present in the training data. Their difference $\Delta$ then directly shows the absolute loss in parsing performance on each section, as if it were an out of domain section. Since in-domain and out-of-domain numbers can be compared, the fact that some sections have a higher overall parsing performance than others does not affect the results. The sections are sorted by $\Delta$LAS.}
\label{tab:tdtinout}
\end{table*}

Finally, in Table~\ref{tab:tdtinout} we compare the in-domain and out-of-domain parsing performance across all sections in the Finnish-TDT corpus by reporting the evaluation performance for both in-domain model where the corresponding section is present in the training data, as well as out-of-domain model, where the section is removed from the training data. In this way we are able to estimate the pure out-of-domain parsing effect, removing the effect of some domains being naturally more difficult to parse than others. While many of the sections do not express substantial differences between the in-domain and out-of-domain performance, especially the legal domain significantly suffers in the out-of-domain setting. Interestingly, when comparing the in-domain performance between different sections, the legal domain receives the second best LAS performance, suggesting the section not being particularly difficult in general but likely the legal text significantly standing out from the other data sources included in the corpus.

\section{Conclusion}

In this work, we introduced a dedicated out-of-domain, manually annotated test set for UD Finnish parser evaluation including data from five distinct text sources previously absent from the UD Finnish treebanks. The selection mirrors practical use cases seen for Finnish dependency parsing in the academia as well as in the industry. In terms of its size, this test set is comparable to other test sets in UD, with its 19,382 syntactic words being considerably above the median UD test set size. Our parsing experiments on this dataset demonstrate that, indeed, syntactic parsing performance can substantially degrade on several domains and the OOD test set now allows us to quantify the effect. On the other hand, we were also able to establish that the effect is at its strongest specifically when measuring the accuracy of the syntactic tree (LAS metric) and is notably less pronounced on the tagging and lemmatization tasks, which have a number of applications in their own right.

Together, these parsing experiments and the Finnish-OOD test set comprise the broadest evaluation of a Finnish state of the art syntactic parser across numerous domains, giving valuable knowledge for all applying the parser outside its training domain in various real-life applications. The new Finnish-OOD treebank is available through the official data releases of the Universal Dependencies framework.

\section{Acknowledgements}

We would like to thank Akseli Leino, Hans Moen and Henry Suhonen for their help with handling the medical terminology. Computational resources were provided by CSC -- the Finnish IT Center for Science, and the research was supported by the Academy of Finland.

\section{Bibliographical References} 

\bibliographystyle{lrec2022-bib}
\bibliography{lrec2022}





\section{Language Resource References}
\label{lr:ref}
\bibliographystylelanguageresource{lrec2022-bib}
\bibliographylanguageresource{languageresource}

\end{document}